\documentclass[runningheads]{llncs}
\usepackage[T1]{fontenc}
\usepackage{graphicx,verbatim}
\usepackage{amssymb}
\usepackage{amsmath}
\usepackage{booktabs}
\usepackage{comment}

\usepackage{array}
\newcolumntype{L}[1]{>{\raggedright\arraybackslash}p{#1}}
\newcolumntype{C}[1]{>{\centering\arraybackslash}p{#1}}
\newcolumntype{R}[1]{>{\raggedleft\arraybackslash}p{#1}}
\usepackage{multirow}

\begin{document}
\title{Histopathology Image Normalization via Latent Manifold Compaction}
\titlerunning{Latent Manifold Compaction}
%
\author{Xiaolong Zhang\inst{1} \and
Jianwei Zhang\inst{2} \and
Selim Sevim\inst{1} \and
Emek Demir\inst{1} \and
Ece Eksi\inst{1} \and
Xubo Song\inst{1}\thanks{Corresponding author. \email{songx@ohsu.edu}}}
\authorrunning{F. Author et al.}
%
\institute{
Knight Cancer Institute, Oregon Health and Science University, Oregon, USA\\
\and
Brenden-Colson Center for Pancreatic Care, Oregon Health and Science University, Oregon, USA
}

\maketitle              

\begin{abstract}


Batch effects arising from technical variations in histopathology staining protocols, scanners, and acquisition pipelines pose a persistent challenge for computational pathology, hindering cross-batch generalization and limiting reliable deployment of models across clinical sites. In this work, we introduce \textbf{Latent Manifold Compaction (LMC)}, an unsupervised representation learning framework that performs image harmonization by learning batch-invariant embeddings from \emph{a single source dataset} through explicit compaction of stain-induced latent manifolds. This allows LMC to generalize to target domain data unseen during training. Evaluated on three challenging public and in-house benchmarks, LMC substantially reduces batch-induced separations across multiple datasets and consistently outperforms state-of-the-art normalization methods in downstream cross-batch classification and detection tasks, enabling superior generalization.


\end{abstract}

\section{Introduction}

Digitized histopathology has become integral to modern clinical workflows and enables machine learning models for disease diagnosis, grading, and prognosis. However, these models often fail to generalize across cohorts, research projects, and institutions due to \textit{batch effects}: systematic variability in slide appearance and feature distributions caused by non-biological technical factors such as tissue processing, staining protocols, and scanner configurations~\cite{Gindra2025ImageAU,kothari2013removing}. Such variability introduces spurious correlations that degrade model performance on unseen data and hinder the integration of multi-source datasets, posing a major barrier to deploying reliable pathology AI systems.

To mitigate batch effects in hematoxylin and eosin (H\&E) histopathology images, a wide range of normalization techniques have been proposed. Classical stain normalization methods align pixel-level color statistics across images~\cite{reinhard2002color,macenko2009method}. While computationally efficient, these approaches primarily adjust global color distributions and may suppress subtle but important biological signals. More recent deep learning approaches—including adversarial style transfer~\cite{shaban2019staingan,kang2021stainnet,zhu2017unpaired}, contrastive stain normalization~\cite{ke2021contrastive,gutierrez2022staincut}, stain-aware domain adaptation~\cite{lafarge2019learning,tiard2022stain,otalora2019staining}, and diffusion-based methods~\cite{shen2023staindiff,jeong2023stain,jewsbury2024stainfuser}—provide greater modeling capacity and flexibility for correcting stain-related variability.

However, most existing methods require access to target-domain data and, in many cases, additional label supervision, which is often infeasible in real-world clinical settings. Unlike natural image deployment scenarios, adapting models with target-site pathology data is constrained by privacy regulations, annotation costs, and institutional data-sharing barriers. Moreover, pre-analytical and preprocessing workflows (such as staining protocols and scanner calibration) are often opaque or poorly documented, making technical variability difficult to model explicitly. As a result, models trained on a single dataset often degrade when deployed \textit{as is}, limiting the aggregation and reuse of multi-source datasets~\cite{aubreville2021quantifying,howard2021impact,tosta2019computational}. More fundamentally, many existing approaches operate at the image level, focusing on harmonizing visual appearance rather than addressing variability in the representation space used by predictive models. As a result, residual batch effects may persist in learned embeddings even after visual normalization. In addition, many methods are tightly coupled to specific architectures or downstream tasks, limiting their use as general-purpose preprocessing modules. Recent studies further show that pathology foundation models remain sensitive to site-specific batch effects, leading to biased cross-batch predictions even with large-scale pretrained encoders~\cite{de2025current}. These limitations highlight the need for approaches that explicitly mitigate batch effects in representation space with strong single source generalizability.

In this work, we present an H\&E normalization method that directly addresses batch effects in learned latent space and enables single-source generalization. We observe that non-biological variations in H\&E images manifest as global intensity shifts in H and E stains individually without changing morphology. Given a single source batch, varying H and E intensities of each image can induce a local 2D manifold in the high dimensional latent space, which captures all possible unseen stain variations corresponding to the same underlying tissue content. Explicitly compacting this manifold to a single point will enforce all varied versions of an image to produce the same embedding, and thus achieving generalized stain-invariant representation while preserving biologically meaningful structure. Therefore, we propose \textbf{Latent Manifold Compaction (LMC)}, an unsupervised batch-invariant representation learning framework using data from a single source batch. LMC leverages principled H\&E stain space augmentations to generate the manifold that reflects real world staining variability and employs a contrastive objective to compact stain variant views into a unified invariant latent representation. Once trained, the resulting encoder can be combined with a task specific prediction head learned on source data and deployed directly on unseen datasets, enabling reliable cross-batch prediction without requiring access to target domain data. We evaluate LMC on three benchmarks that emphasize cross-batch generalization~\cite{aubreville2021quantifying,litjens20181399,shaban2019staingan}, where models are trained exclusively on a single source dataset and tested on distinct target datasets acquired under different staining conditions. LMC consistently reduces batch-induced separation and achieves superior cross-batch performance compared to state-of-the-art normalization methods.

\section{Method}
\label{sec:method}

By design, stain normalization is achieved by first generating the manifold for each image by varying H and E intensities which captures all possible stain variations corresponding to the same underlying tissue content. Explicitly compacting this manifold to a single point can then produce generalized stain-invariant representations while preserving biologically meaningful structure.

\subsection{Stain-Induced Manifold Generation}\label{sec:lat-manifolds}

\begin{figure}
\centering
\includegraphics[width=0.95\textwidth]{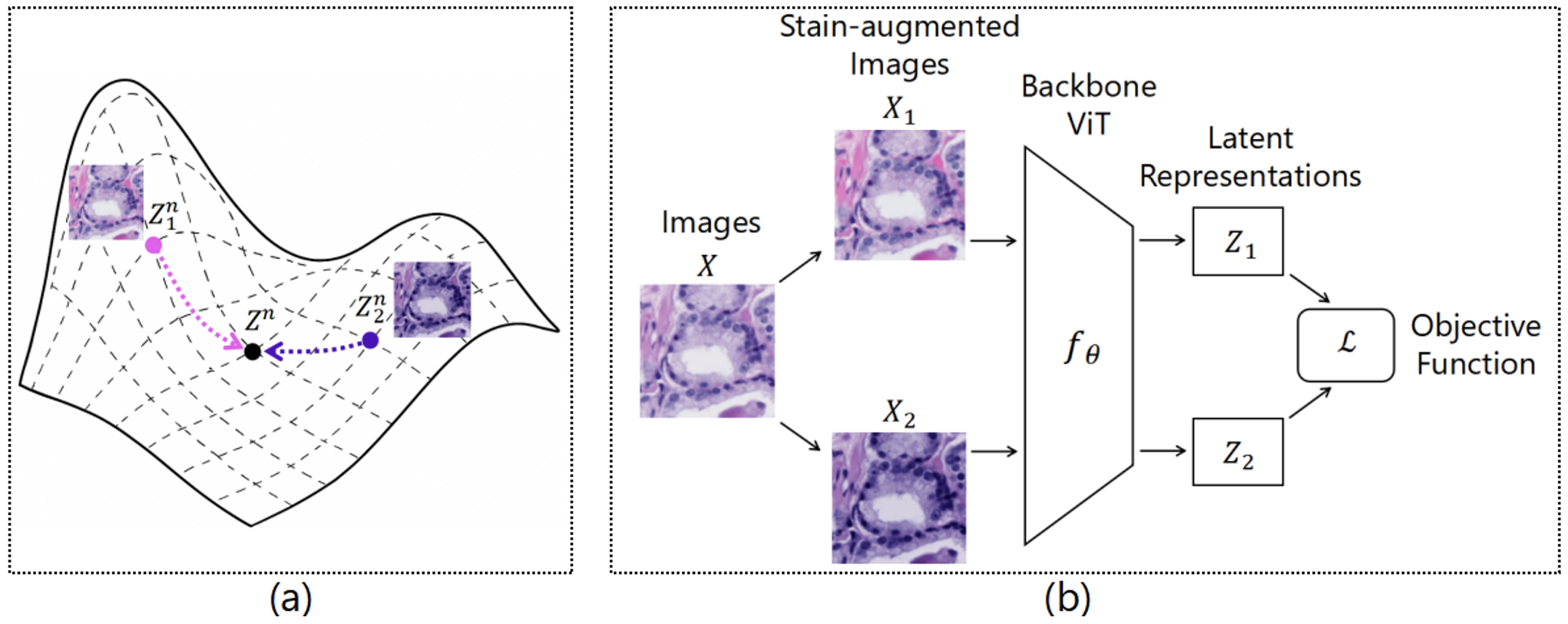}
\caption{(a) Manifold generated by varying H and E intensities of each image induces a local 2D manifold in latent space. All points on the manifold are compacted to a single point $Z^n$, which is a stain-invariant representation that reduces batch effects while preserving biological structure. (b) LMC applies H\&E stain-space augmentations, encodes paired views with a Vision Transformer (ViT) backbone, and enforces batch-invariant representations using a contrastive loss.}

\label{fig:lmc}
\end{figure}

To generate the stain-induced manifold for a histopathology image, we produce multiple stain-augmented views of the image, as illustrated in Figure~\ref{fig:lmc}(a). Specifically, a H\&E image in RGB space is first deconvolved into the H and E channels by singular value decomposition (SVD), which is performed on the optical density (OD) matrix of the image to identify the principal eigenvectors corresponding to hematoxylin (H) and eosin (E), thereby defining the H\&E space. Then the H and E components are scaled with parameters $\alpha_{H}$ and $\alpha_{E}$ respectively, to reflect the stain variation. We choose $\alpha_{H}$ and $\alpha_{E}$ from a uniform range of [0.5,2.0] that empirically cover the range of stain variation in the real world to populate the manifold. The resulting augmented image is obtained by reconstructing the RGB from the perturbed H\&E space. These controlled stain perturbations to histopathology images reflect variability observed in practice and produce diverse yet biologically consistent stain variants.

\subsection{Manifold Compaction in Latent Space}\label{sec:contrastive-objective}

The principled objective is to compact the stain-induced latent manifold toward a single semantically meaningful representation, as shown in Figure~\ref{fig:lmc}(a). This is done by training an encoder such that the embeddings of any pairs of points on the manifold to be close. Given an input image $x$, we denote the two independently stain-augmented views as $x_1$ and $x_2$, as shown in Figure~\ref{fig:lmc}(b). An encoder $f_\theta(\cdot)$ maps each view into a latent representation:
\begin{equation}
z_1=f_{\theta}(x_1), \quad z_2=f_{\theta}(x_2).
\end{equation}

To efficiently compact stain-induced latent manifolds while avoiding reliance on large batch sizes or memory banks~\cite{oord2018representation,chen2020simple,he2020momentum}, training instability handling~\cite{bardes2021vicreg}, and carefully tuned regularization~\cite{grill2020bootstrap,chen2021exploring}, we employ a correlation-based contrastive objective that enforces invariance between paired representations while reducing redundancy across embedding dimensions~\cite{zbontar2021barlow}. Specifically, for a mini-batch of paired representations, we compute the cross-correlation matrix between corresponding latent dimensions and optimize an objective that (i) encourages strong alignment along the diagonal and (ii) suppresses off-diagonal correlations. The resulting objective function is expressed as:
\begin{equation}\label{eq:total-loss}
\mathcal{L}
\;\triangleq\;
\sum_i \left(1 - C_{ii}\right)^2
\;+\;
\lambda \sum_i \sum_{j \neq i} C_{ij}^2,
\end{equation}

\noindent where $\lambda$ is a hyperparameter that balances the two loss terms and is set to 0.005 based on empirical tuning. $C_{ij}$ is the cross-correlation matrix between paired latent representations $z_1$ and $z_2$ (in Figure~\ref{fig:lmc}(b)) for one batch of training inputs:

\begin{equation}
C_{ij}
\;\triangleq\;
\frac{\sum_b z^{b}_{1,i}\, z^{b}_{2,j}}
{\sqrt{\sum_b \left(z^{b}_{1,i}\right)^2}\;
 \sqrt{\sum_b \left(z^{b}_{2,j}\right)^2}},
\end{equation}

\noindent where $b$ indexes samples in a training batch and $i,j$ index the dimensions of latent representations.

The first term of Eq.~\ref{eq:total-loss} enforces \emph{invariance} by driving the diagonal elements of the cross-correlation matrix toward one, ensuring that paired stain-variant views map to highly aligned embeddings. The second term promotes \emph{redundancy reduction} by suppressing off-diagonal correlations, encouraging different latent dimensions to encode complementary information. Together, these effects compact stain induced latent manifolds into stable, low redundancy representations that preserve biological semantics while removing stain dependent variation.

Notably, this formulation achieves invariance without explicit negative samples, which is desirable in histopathology where distinct tissue patches may exhibit highly similar morphology~\cite{kothari2012biological}. By avoiding instance discrimination based repulsion~\cite{oord2018representation,chen2020simple,he2020momentum}, LMC maintains meaningful semantic proximity for similar image patches while enforcing stain invariance.


\section{Experiments}

We evaluate LMC on three clinically relevant benchmarks that exhibit substantial batch effects due to differences in staining protocols, scanners, and acquisition settings. In all experiments, the encoder is trained \textit{exclusively on a single source dataset} and evaluated on one or more unseen target datasets, reflecting realistic deployment scenarios. 

\subsection{Experimental Settings}\label{sec:exp-settings}

\paragraph{ViT Backbone and Training.} A lightweight ViT with 5.5 million parameters is adopted as the LMC encoder, which contains 12 transformer encoder blocks, 3 multi-attention heads, with residual connections and an embedding dimension of 192. During training, we adopt a staged learning-rate schedule consisting of an initial warm-up phase, followed by a stable training stage with learning rate 1e-4, and a final annealing phase in which the learning rate is gradually reduced to 1e-7. The model is optimized using the AdamW optimizer with a weight decay of 0.01 and a batch size of 32 on an NVIDIA A40. Once trained, the encoder can be directly used as a feature extractor for downstream tasks on unseen datasets.

\paragraph{Baselines.} We compare LMC against \emph{Unnormalized} case, classical \emph{Macenko} stain normalization~\cite{macenko2009method}, and the most recent diffusion-based normalization method \emph{StainFuser}~\cite{jewsbury2024stainfuser} as baselines in our evaluations.

\subsection{Tumor Metastasis Classification}\label{sec:exp-tumor}

\paragraph{Dataset and Setup.} We use the public Camelyon16 dataset, which contains annotated H\&E whole-slide images (WSIs) from two institutions: 249 WSIs from Radboud University Medical Center (RAD) and 150 WSIs from University Medical Center Utrecht (UNI)~\cite{litjens20181399}. The task is to classify metastatic (tumor) tissues from non-metastatic (normal) breast cancer images across different datasets. The LMC encoder is trained using 353,858 unlabeled 256×256 patches extracted from RAD WSIs only. For downstream evaluation, a classifier is trained on publicly available labeled RAD patches and tested directly on UNI patches~\cite{shaban2019staingan}, without any access to UNI data during training.

\begin{figure}
\centering
\includegraphics[width=0.95\textwidth]{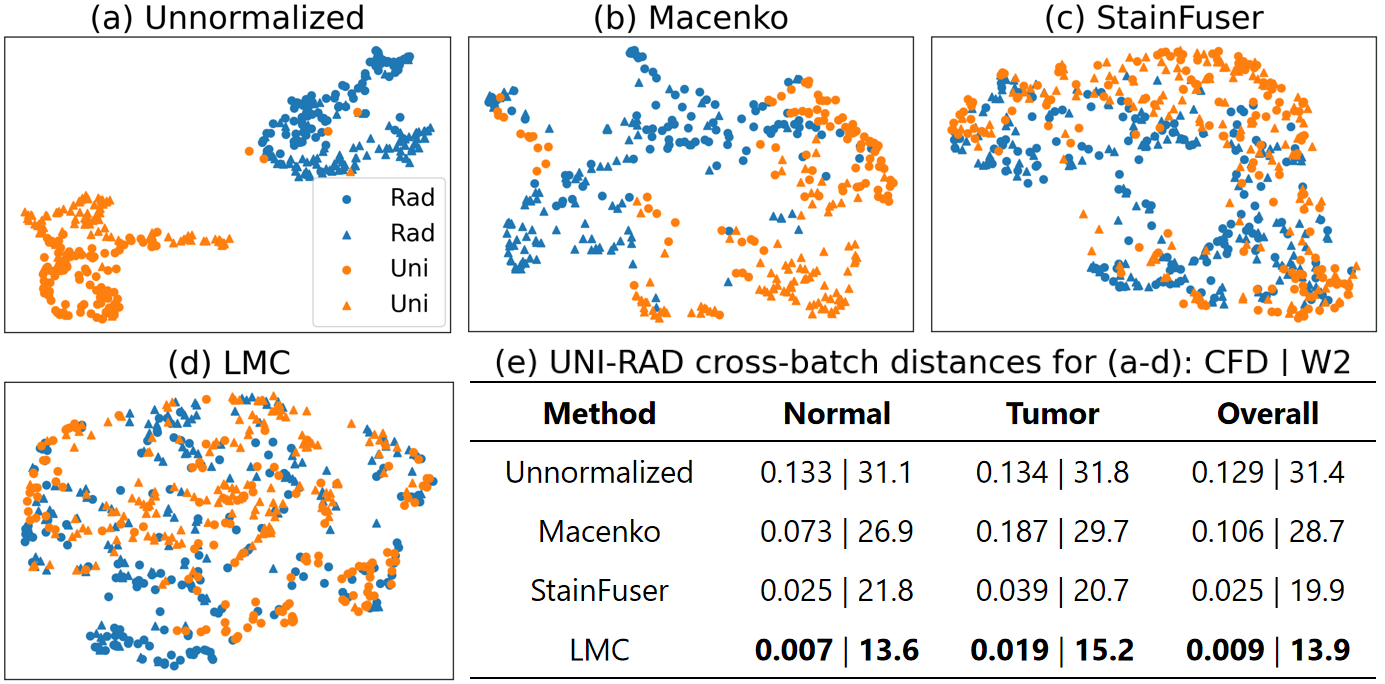}
\caption{Camelyon16 visualization and quantification of batch separation results. (a-d) Qualitative UMAPs. (e) Quantitative batch separations measured by CFD and W2. \raisebox{0.25ex}{\scalebox{0.75}{$\bigcirc$}}-Normal, \raisebox{-0.05ex}{\scalebox{0.9}{$\triangle$}}-Tumor.}
\label{fig:camelyon-umap}
\end{figure}

\paragraph{Results.} Figure~\ref{fig:camelyon-umap} presents UMAP visualizations of latent representations under different normalization strategies, together with quantitative measurements of residual batch separation between RAD and UNI using the Wasserstein-2 (W2) distance and Cross-Fusion Distance (CFD)~\cite{zhang2026cross}, which quantify batch separation in latent space and provide interpretable measures of batch effects. All latent representations corresponding to compared methods are extracted using pathological foundation model Virchow~\cite{vorontsov2024foundation}. Without normalization, RAD and UNI form clearly separated clusters, indicating pronounced batch effects. Macenko normalization moderately reduces this separation while preserving intrinsic biological meanings, with normal (circle symbols) and tumor (triangle symbols) samples occupying distinct regions in the UMAP. StainFuser further decreases overall batch separation; however, the distinction between tumor and normal classes becomes less pronounced. 

In contrast, LMC produces substantially overlapping latent distributions with minimal residual separation, reflected by the lowest CFD and W2 values for both normal and tumor classes as well as overall. Importantly, LMC maintains clear distinction between normal and tumor classes while effectively aligning corresponding classes across sites. These findings indicate that LMC most effectively mitigates batch-induced separations while preserving class-discriminative biological structure, thereby promoting stronger and more meaningful cross-batch generalization among the compared normalization methods.

\begin{figure}
\centering
\includegraphics[width=0.8\textwidth]{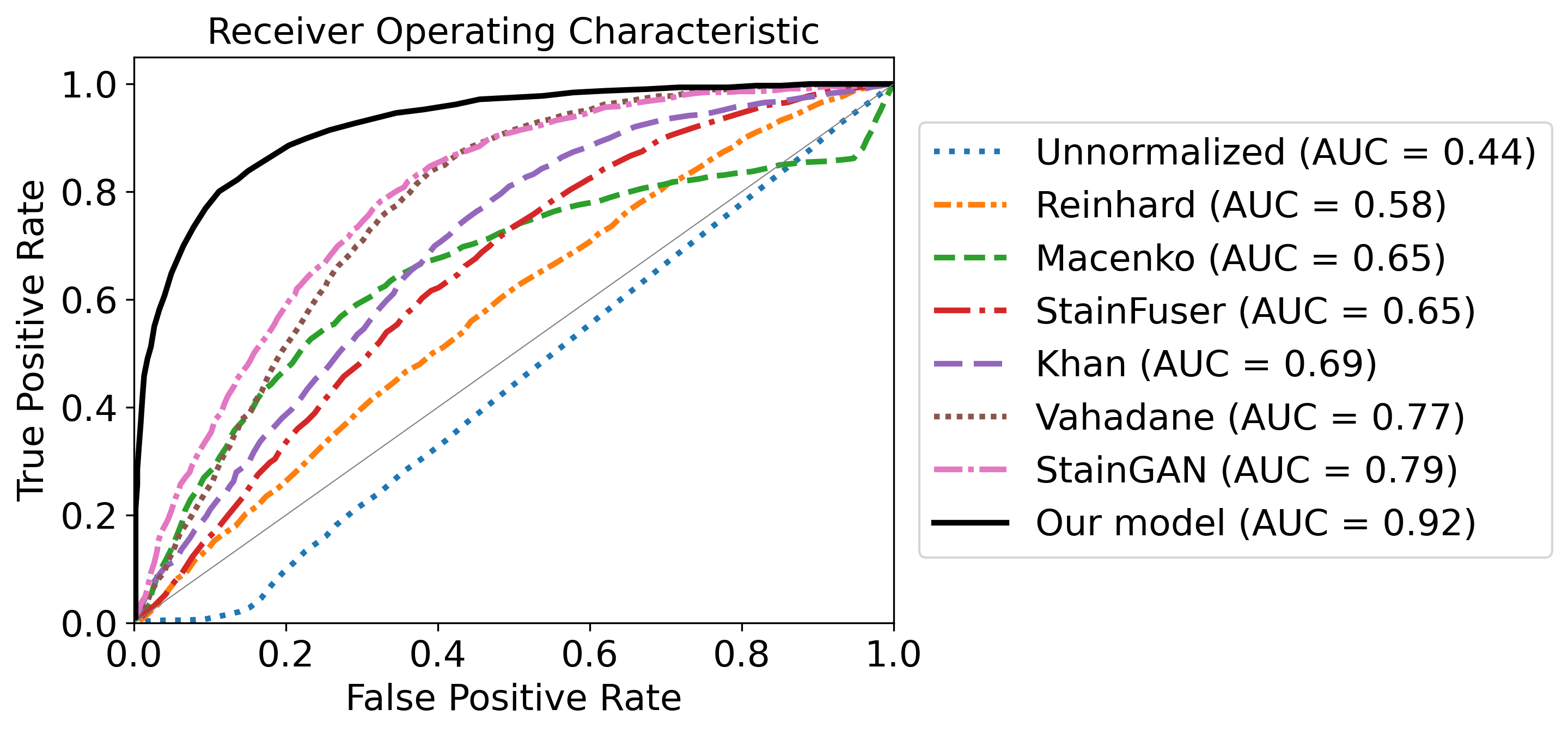}
\caption{Camelyon16 RAD-trained Model Cross-batch Test (on UNI): ROC and AUC. Our model (solid black line) achieves the best AUC among all compared methods.}
\label{fig:camelyon_auc_roc}
\end{figure}

Cross-batch classification ROC curves and AUC results are shown in Figure~\ref{fig:camelyon_auc_roc}, together with results from previously reported methods in the literature~\cite{shaban2019staingan}. LMC achieves the highest cross-dataset AUC, outperforming both classical and deep learning–based normalization approaches and indicating improved decision-boundary transfer under realistic deployment conditions.

\subsection{Prostate Multi-class Gleason Grading}\label{sec:exp-prostate}
\paragraph{Dataset and Setup.} We evaluate multi-class generalization on in-house prostate cancer datasets collected under two different clinical workflows: a needle biopsy bio-repository (BR) dataset comprising 102 patient WSIs and a prostatectomy bio-library (BL) dataset comprising 18 patient WSIs. Both datasets are manually annotated by expert pathologists, with labeled prostate tissue classes including benign, Gleason grade 3 (G3), and multiple Gleason grade 4 (G4) subtypes, with class frequencies ranging from approximately 44.3\% for benign to 0.77\% for rare G4-glomeruloid subtype patterns. 

The LMC encoder is trained on 237,940 unlabeled 256×256 patches extracted exclusively from BR WSIs. For downstream evaluation, a multi-class classifier is trained using BR's 18,043 labeled tissue patches, without access to BL data during training, and evaluated directly on BL's 12,112 labeled patches.

\paragraph{Results.} Cross-batch multi-class classification results are presented in Table~\ref{tab:prostate-cross-acc}. While all normalization methods improve performance relative to the unnormalized baseline, LMC achieves the highest overall accuracy and consistently outperforms all competing methods across multiple classes and Gleason patterns. These results demonstrate that LMC remains effective even in the presence of substantial differences in tissue preparation and staining protocols between datasets. 

\begin{table}[htbp]
\centering
\caption{BR-trained Model Cross-batch Test (on BL) Accuracy (by Class)}
\label{tab:prostate-cross-acc}
\begin{tabular}{L{3.8cm} C{2.25cm} C{1.8cm} C{1.5cm} C{1.5cm}}
\toprule
\textbf{Classes}
& \textbf{Unnormalized}
& \textbf{Macenko}
& \textbf{StainFuser}
& \textbf{LMC} \\
\midrule
G3  & 0.1\% & 57.8\% & 2.5\% & \textbf{58.0\%} \\
G4-cribriform subtype   & 0.0\% & 3.6\% & 0.0\% & \textbf{30.6\%} \\
G4-poorly formed subtype & \textbf{99.9}\% & 37.3\% & 79.7\% & 54.3\% \\
G4-fused subtype  & 0.0\% & 0.0\% & 1.4\% & \textbf{17.4\%} \\
G4-glomeruloid subtype   & 0.0\% & 0.0\% & 0.0\% & \textbf{5.6\%} \\
Benign & 0.9\% & 12.6\% & \textbf{65.4}\% & 44.3\% \\
Overall & 10.8\% & 25.4\% & 29.1\% & \textbf{45.7\%} \\
\bottomrule
\end{tabular}
\end{table}

\subsection{Mitotic Figure Detection}\label{sec:exp-mitotic}

\paragraph{Dataset and Setup.} We use Mitosis Domain Generalization (MIDOG) 2021 challenge dataset~\cite{aubreville2023mitosis}, from which we use 150 annotated WSIs: 50 acquired with an Aperio ScanScope CS2 (ACS), 50 with a Hamamatsu S360 (HS), and 50 with a Hamamatsu XR (HXR) scanner. The LMC encoder is trained using 206,752 unlabeled 256×256 patches extracted from ACS WSIs only. A classifier is then trained on labeled ACS (665 mitotic and 893 hard negative) patches, and evaluated on labeled HS (562 mitotic and 1,029 hard negative) and HXR (438 mitotic and 704 hard negative) patches.

\paragraph{Results.} Table~\ref{tab:midog-cross-acc} reports F1 scores of cross-batch test on HS and HXR for mitosis detection. All normalization methods improve performance relative to the unnormalized baseline, though to varying extents. Macenko and the most recent method StainFuser achieve moderate gains, with average F1 scores of 0.482 and 0.439, respectively, whereas LMC achieves the highest F1 scores on both HS (0.600) and HXR (0.651), resulting in the best average performance (0.626). These results demonstrate substantially stronger cross-batch generalization of LMC for mitosis detection under heterogeneous acquisition conditions commonly encountered in clinical practice.

\begin{table}[htbp]
\centering
\caption{MIDOG21 ACS-trained Model Cross-batch Test (on HS \& HXR) F1 Scores}
\label{tab:midog-cross-acc}
\begin{tabular}{L{2.5cm} C{2.5cm} C{2.5cm} C{2.5cm}}
\toprule
\textbf{Method}
& HS 
& HXR 
& Average \\
\midrule
Unnormalized & 0.294 & 0.443 & 0.369 \\
Macenko      & 0.453 & 0.511 & 0.482 \\
StainFuser   & 0.418 & 0.460 & 0.439 \\
LMC         & \textbf{0.600} & \textbf{0.651} & \textbf{0.626} \\
\bottomrule
\end{tabular}
\end{table}

\section{Conclusion}

We proposed Latent Manifold Compaction (LMC), an unsupervised representation learning framework that mitigates batch effects in computational pathology by explicitly compacting stain-induced manifolds in latent space. Trained on a single source dataset, LMC learns batch-invariant representations that generalize effectively to unseen datasets without requiring access to target-domain data. This yields a task‑agnostic, patch‑wise, feature‑level harmonization mechanism that is naturally compatible with a wide range of downstream models and training regimes. Evaluations on three public and in-house benchmarks demonstrate that LMC consistently reduces batch-induced separation and outperforms both classical and deep learning–based stain normalization methods in cross-batch tasks. Importantly, these evaluations reflect realistic deployment settings in which models trained on a single source site are applied directly to previously unseen target sites. Future work will explore extending LMC beyond H\&E histopathology to other imaging modalities, scaling to larger foundation models, and validating its effectiveness across broader clinical tasks.



%
%
\bibliographystyle{splncs04}
\bibliography{ref}
\end{document}